\documentclass[letterpaper, 10 pt, conference]{ieeeconf}  
\IEEEoverridecommandlockouts                           
 \usepackage{graphics} 
 \usepackage{graphicx}
 \usepackage{epsfig} 
\usepackage{amsmath}
 \usepackage{booktabs}
 \usepackage{array}
 \usepackage{float}
\usepackage[space]{cite}    
\usepackage[hidelinks]{hyperref}
\usepackage[capitalize]{cleveref}  

\title{\LARGE \bf
Whisker-Inspired Tactile Sensing:\\
A Sim2Real Approach for Precise Underwater Contact Tracking
}

\usepackage[dvipsnames]{xcolor}

\author{Hao Li$^{*}$, \space Chengyi Xing$^{*}$, \space Saad Khan, \space Miaoya Zhong, \space Mark R. Cutkosky
\thanks{*Equal contribution}
\thanks{The authors are with Stanford University, USA {\tt \{li2053, chengyix, saadkh, miaoya, cutkosky\} @stanford.edu}}}

\begin{document}
\maketitle
    \thispagestyle{empty}
\pagestyle{empty}
\begin{abstract}
Aquatic mammals, such as pinnipeds, utilize their whiskers to detect and discriminate objects and analyze water movements, inspiring the development of robotic whiskers for sensing contacts, surfaces, and water flows. We present the design and application of underwater whisker sensors based on Fiber Bragg Grating (FBG) technology. These passive whiskers are mounted along the robot’s exterior to sense its surroundings through light, non-intrusive contacts. For contact tracking, we employ a sim-to-real learning framework, which involves extensive data collection in simulation followed by a sim-to-real calibration process to transfer the model trained in simulation to the real world. Experiments with whiskers immersed in water indicate that our approach can track contact points with an accuracy of $<2$\,mm, without requiring precise robot proprioception. We demonstrate that the approach also generalizes to unseen objects.
\end{abstract}

\section{INTRODUCTION}

Whiskers, or vibrissae, are vital sensory organs that enable animals to acquire crucial information about their environment, especially in situations where vision is compromised. In low-light or cluttered environments, terrestrial animals like rats and cats use their whiskers to detect and navigate around obstacles, effectively sensing nearby objects and surfaces \cite{boublil2021mechanosensory}. Similarly, underwater animals such as pinnipeds, which have limited visual acuity in turbid water, rely on their sensitive whiskers to detect and differentiate objects as well as analyze water movements \cite{hanke2013hydrodynamic, murphy2017seal, eberhardt2016development}, aiding them in prey detection and navigation.

Underwater robots similarly can benefit from whiskers in cluttered scenarios. Robots such as OceanOne sometimes operate in murky water and have even become stuck on underwater obstacles \cite{khatib2016ocean}. If cameras are not able to see where the contact occurs, freeing the robot may require external assistance. Recently, commercial counterparts to OceanOne have been developed (e.g. \cite{honda2024rov}) and these too may benefit from contact or proximity sensing for safe operation.

\begin{figure}[t]
      \centering
      \includegraphics[width=0.95\linewidth]{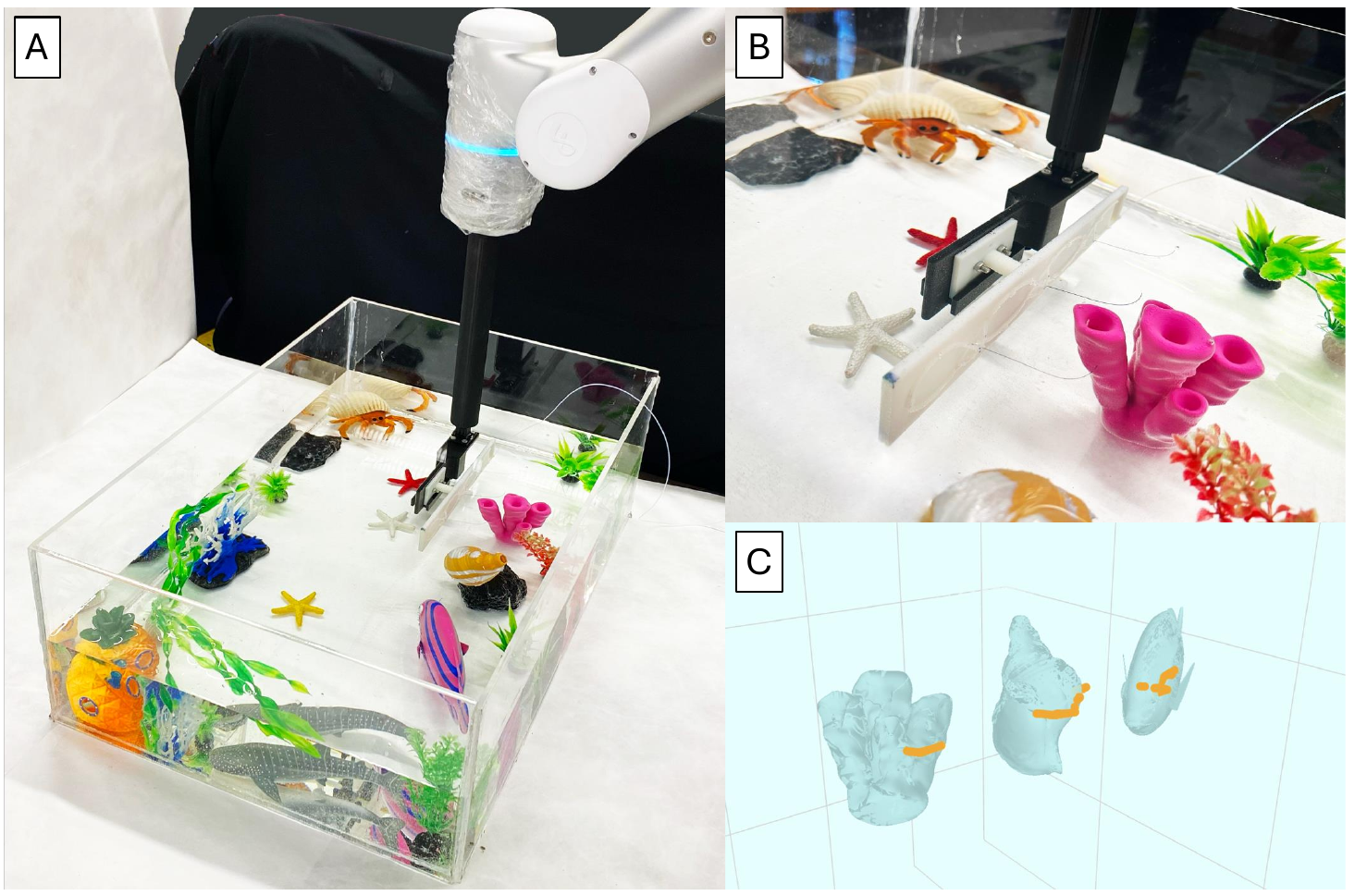}
      \caption{(A) Robot arm instrumented with a whisker sensor array in a crowded fish tank. (B) A close view of the whisker array in water. (C) Predicted contact positions when whiskers sweep over 3 objects.}
      \label{demo}
\end{figure}

Although whiskers are not as widely employed in robotics as other sensing modalities, a number of examples appear in the literature. Some approaches involve active whisking mechanisms with rotational joints and actuators at the whisker base \cite{kaneko19953, kaneko1998active, kim2007biomimetic,prescott2009whisking}, while others utilize passive whiskers for contact sensing and tracking \cite{merker2021vibrissa, huet2017tactile, sofla2023spatial, sofla2024haptic}. Most designs incorporate a thin elastic rod
attached to a one- or two-axis base sensor to detect contacts at the whisker tip or, in more advanced applications, to track continuous contacts along a deflected whisker. Additional discussion on prior related work is included in \cref{sec:related}.

In the present work we report on whiskers designed for underwater operation (\cref{demo}). The transduction technology employs optical fibers with Bragg gratings (FBGs) enbedded within 3D printed structures. The whiskers are Nitinol (TiNi) wires. This choice of technology is motivated by the ultimate application: operation in deep, corrosive salt water with large changes in ambient pressure as the robot moves up and down. Optical fibers with FBGs have demonstrated robustness in harsh environments, including deep oil and gas wells \cite{qiao2017fiber}. Functioning as optical strain gages, they are highly sensitive, able to resolve strains of $10^{-5}$. Strains in the FBGs produce shifts in reflected optical wavelengths, which are read by an optical interrogator. The interrogator can be located as much as 1\,km away, allowing it to reside on a surface boat while the robot operates below. In addition, it is possible to arrange dozens of FBGs along a single  fiber, each with a sightly different nominal wavelength---and sample all of them at 2 kHz rates. 

\begin{figure*}[t!]
      \centering
      \includegraphics[width=0.95\linewidth]{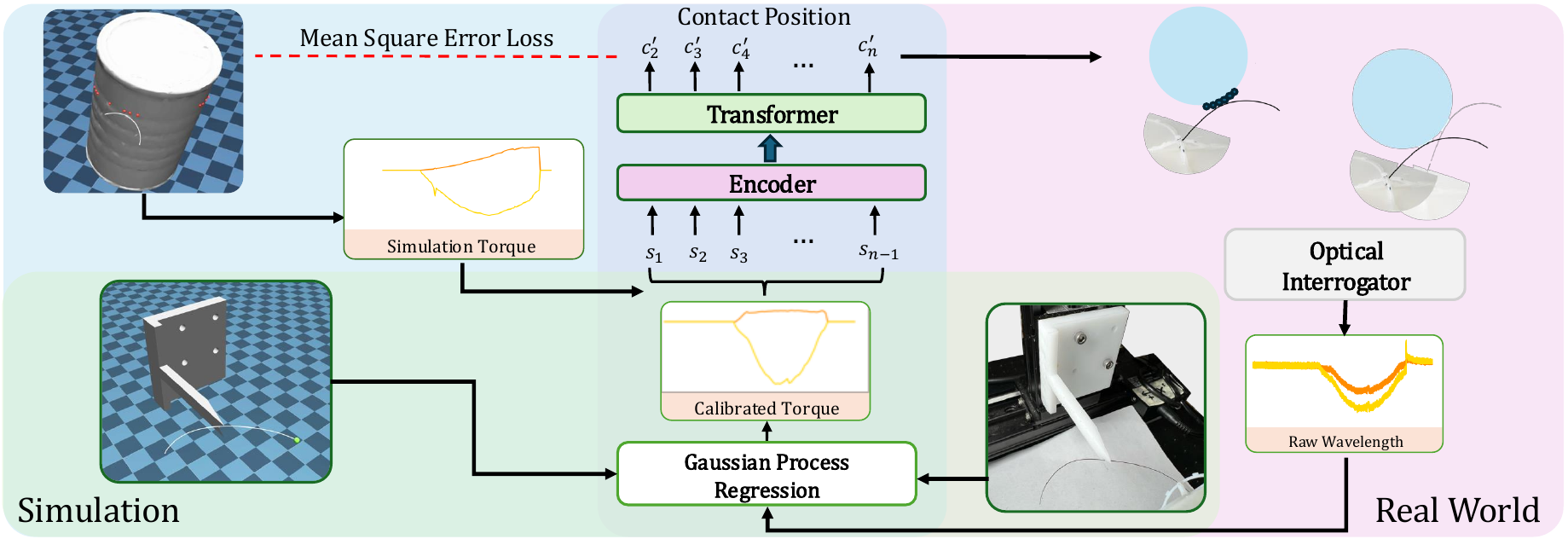}
      \caption{Overview of our sim2real learning framework. Moments computed in simulation are processed and used as inputs for training WhiskerNet, which is optimized using mean squared error (MSE) loss to predict contact locations. A calibration process, leveraging Gaussian Process Regression, maps wavelength data to base moments in simulation and real-world scenarios. The trained WhiskerNet model is then deployed in the real world for accurate contact location prediction.}
      \label{overview}
\end{figure*}

Apart from the process of converting whisker deflections to data, there are a number of issues associated with interpreting the results. In comparison to some previous work, we use pre-curved whiskers to avoid the highly nonlinear buckling behavior that can occur when straight whiskers approach a surface in the axial direction. In addition, we are interested in knowing not only when the tip of a whisker makes contact but also the path of the contact location as a curved whisker brushes over a surface. With this information we can track whisker/surface contacts and ultimately reconstruct surfaces.

\textbf{Contributions}: We introduce a whisker sensor design that uses a single optical fiber with multiple FBGs to record data from a patch of whiskers. Although the use of optical fibers for whiskers or whisker-like sensors is not unprecedented \cite{zhao2017novel, ding2023low, zhang2023novel, glick2024tracking, wang2024biomimetic, fujiwara2024optical}, the design presented here is particularly suited for tracking whisker/surface contacts and for operation in the deep ocean, as it is unaffected by saltwater or high ambient pressure.

In contrast to our previous work \cite{lin2024navigation}, which used Bayesian methods and relied on accurate proprioception data from the robot, we use a deep learning approach to track whisker/object contacts. To meet the requirement for large amounts of data, we employ a sim-to-real framework with the following components: (1) a digital twin in MuJoCo to generate large-scale datasets of whiskers sweeping over various objects from the YCB dataset, (2) a deep learning model based on a transformer-decoder architecture that processes sensor signal history to predict the next contact location, and (3) a calibration method that maps real-world signals to simulation signals using Gaussian Process Regression (GPR). The overall framework is shown in \cref{overview}. This approach does not rely explicitly on robot proprioceptive data and is therefore suitable for use on a mobile robot, which may be subject to drift. 

To validate the effectiveness of our design and methods, we conduct experiments in both simulated and real-world environments, including tests in air and immersed in water. The water tests are conducted using a robot arm to drag a set of whiskers through a shallow water tank; this is a first step toward operation on a mobile robot in deep water. A variety of previously identified and new, unseen objects are evaluated, with our metric measuring the distance from predicted contact points to the nearest object surface. Our experiments demonstrate that our method achieves contact prediction accuracy within 2\,mm for most objects. The main discrepancies arise from imperfect whisker/object tracking as whiskers slide over surfaces with steep vertical angles or abrupt changes in contour. Additionally, we conduct an ablation study on the impact of different movement speeds to show robustness in real-world applications.

\section{Related Work}
\label{sec:related}
\subsection{Robotic Whisker Sensing}
In comparison to the approach of considering only tip-contact for relatively stiff whiskers, some works have considered continuous contacts along complaint whiskers as they brush over surfaces. Early approaches consisted of a flexible beam, a rotation motor, and a torque sensor, to locate contact points using rotational compliance \cite{kaneko19953,kaneko1998active}. Subsequent works were able to discriminate various object shapes by measuring deflection angles and velocities \cite{kim2007biomimetic,merker2021vibrissa}. A recent review of whisker sensing in robotics is provided in \cite{yu2024whisker}.

Other work has employed model-based methods to estimate contact points using a 3-axis force sensor at the base of a compliant whisker in 3D \cite{huet2017tactile, sofla2024haptic, sofla2023spatial}. The solution is not straightforward, however, because isolated torque measurements at a compliant whisker base do not correspond to a unique contact location. 
Additional information can be incorporated for higher accuracy. For example, \cite{lin2022whisker,lin2024navigation} achieved an accuracy of within 2\,mm for contact tracking using a Bayesian filtering method that incorporates the known velocity of the whisker base on a robot arm moving among stationary objects. For a mobile robot, we desire an approach that does not require this information.

Whiskers in water are also useful to detect water currents, for example from nearby moving objects. This is perhaps the most important use of whiskers in pinnipeds, and it has been used in robotic implementations as well \cite{liu2023artificial, wang2024bio, muthuramalingam2019seal}. Some works also use optical fiber-based whiskers to detect flow and vortices. For example, \cite{glick2024tracking} estimate the direction and velocity of flow disturbances by treating the optical fiber as a whisker and sensing the strain applied at the base using one FBG sensor.  
In \cite{wang2024biomimetic} a resin whisker with a shaft embedded with four FBG sensors measures flow velocity. Our work focuses on contact detection with thin, metallic whiskers and low velocities so that water has little affect on whisker curvature or contact locations.

\subsection{One-shot Sim-to-Real Transfer}
The rapid advancement of deep learning has led to a pivotal change in robotics research, steering it toward approaches that require large datasets for training purposes. However, generating such datasets using robots can be prohibitively expensive and time-consuming. To address this challenge, robotic simulators like PyBullet\cite{coumans2016pybullet}, MuJoCo\cite{todorov2012mujoco}, Drake\cite{tedrake2019drake}, and Isaac Sim\cite{nvidia_isaac_sim_2022} have been developed, enabling efficient and cost-effective data collection.

While these simulators are becoming increasingly accurate, they are implemented based on a set of physical models and, therefore, imperfect. To bridge the reality gap, one-shot transfer techniques like domain randomization and system identification are employed \cite{zhao2020sim}, perturbing visual and dynamics parameters (e.g., texture, lighting, camera pose, friction, damping, and geometry) and identifying physical parameters through real-to-sim-to-real calibration to align simulations with reality, thereby improving model transferability. These methods have facilitated the successful transfer of learned models to physical tasks such as navigation\cite{kadian2020sim2real, mitriakov2021reinforcement, traore2019continual, gao2023sonicverse, kaufmann2020deep}, locomotion\cite{lee2020learning, tan2018sim, peng2020learning, hanna2017grounded}, and manipulation\cite{akkaya2019solving, andrychowicz2020learning, rao2020rl, matl2020inferring}.

We follow a similar methodology by simulating whiskers in MuJoCo. We align the whisker shape in simulation to real world measurements, and brush whiskers against various objects in simulation with different trajectories and velocities, generating a large dataset for training purposes. While it is common to align simulation data with real-world data, we instead calibrate measured shifts in FBG wavelengths to match simulated torque (bending moment) 
signals, as described in Section \ref{sec:whisker_sim_to_real}. This ``real-to-sim'' calibration allows us to deploy our model for navigating real-world scenarios, although it limits performance to the fidelity of the simulator’s dynamics.

\begin{figure}[t!]
      \centering
      \includegraphics[width=0.9\linewidth]{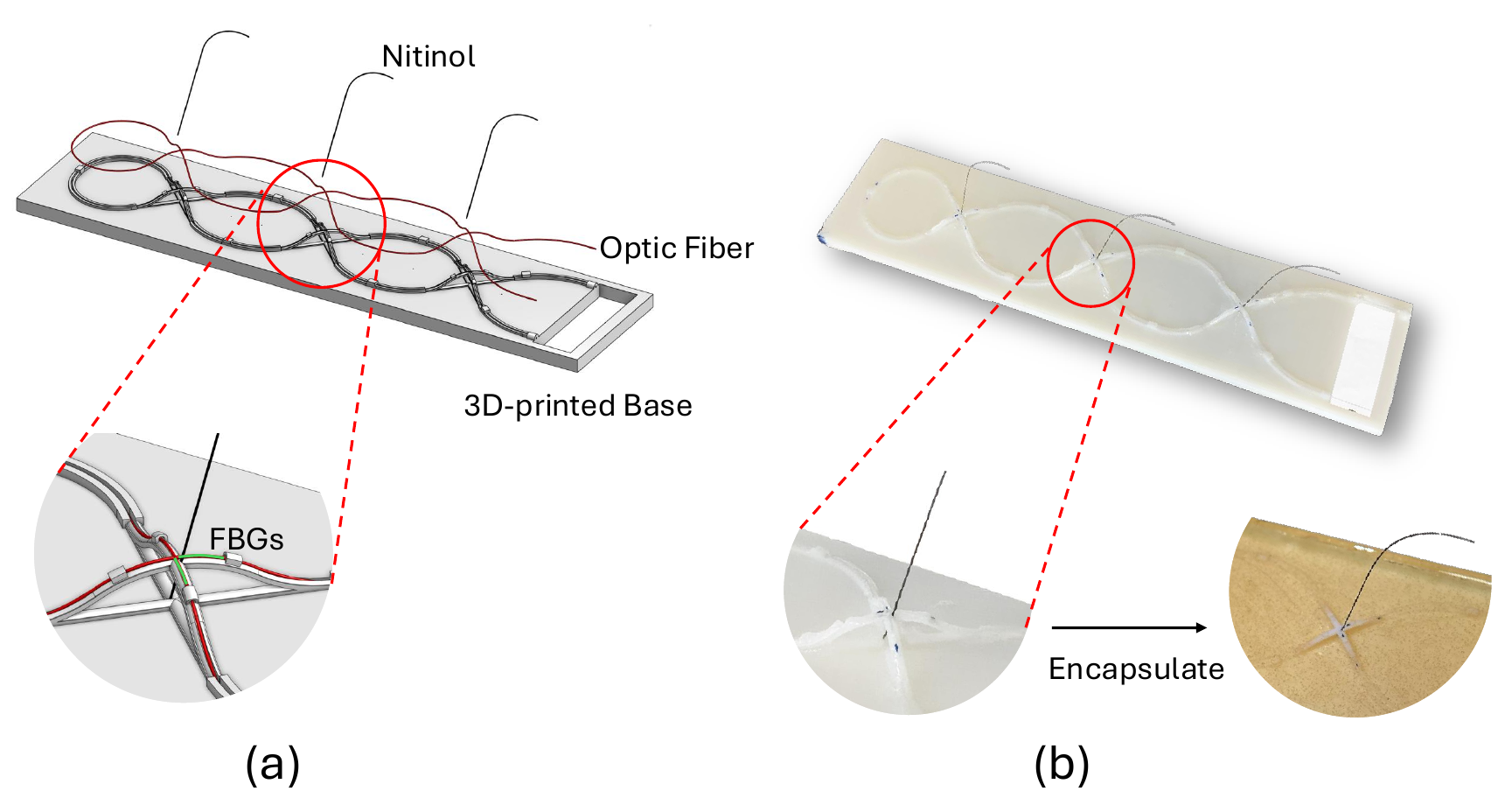}
       \caption{(a) Schematic of the whisker sensor design and components. (b) Image of the fabricated and encapsulated whisker sensor.}
      \label{design}
\end{figure}

\begin{figure}[t!]
      \centering
      \includegraphics[width=0.9\linewidth]{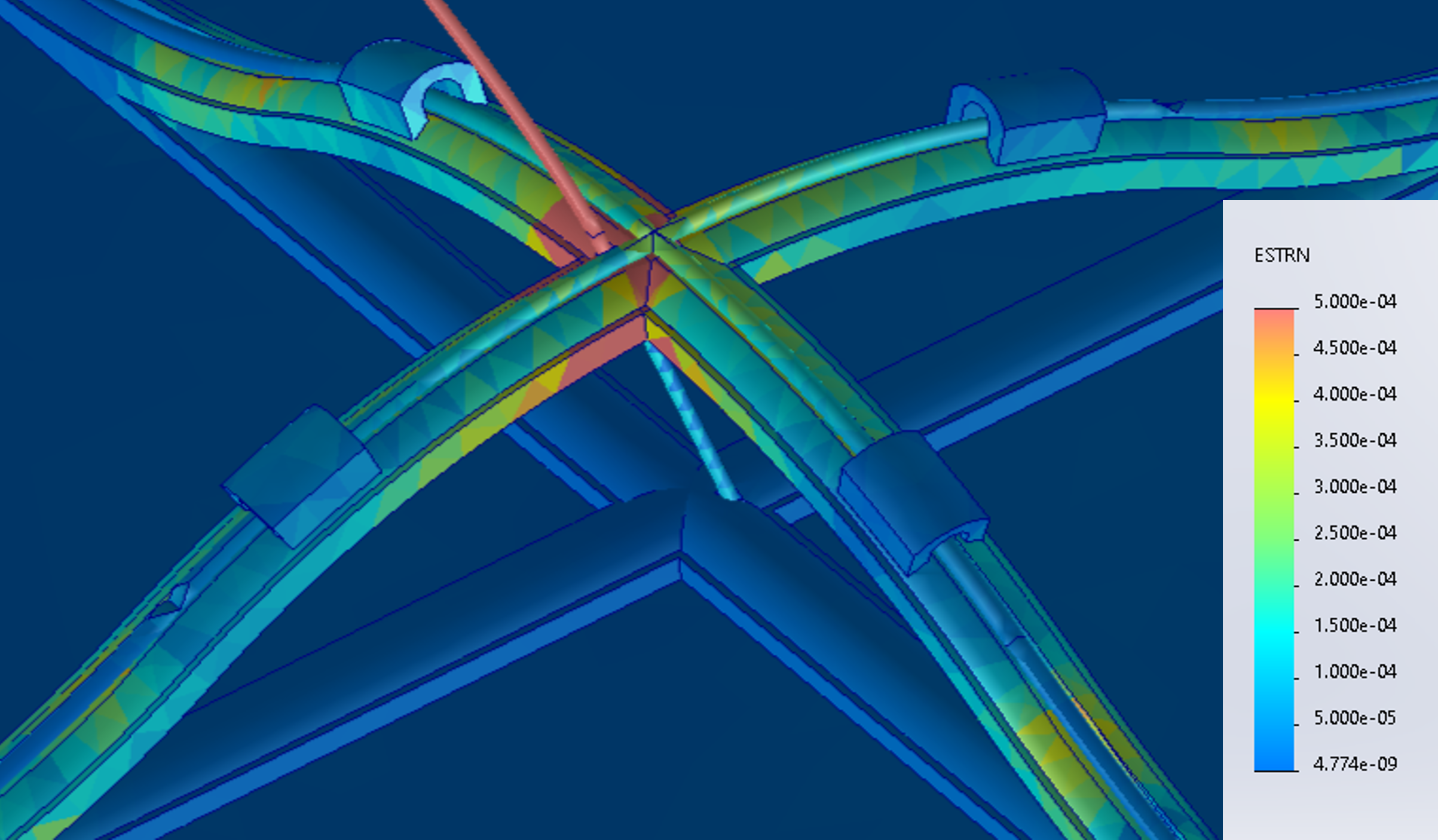}
      \caption{Finite element analysis of the bridge structure in the whisker sensor design, illustrating the distribution of equivalent strain across the structure. The color gradient represents strain magnitude, with red denoting regions of higher strain.}
      \label{fea}
\end{figure}

\section{Sensor Design and Fabrication}

Fig. \ref{design} shows an overview of our underwater sensor design and components. The whiskers consist of thin, precurved, super-elastic ASTM F2063 nitinol wires, 0.3\,mm diameter, with a distal 20\,mm arc radius, and a 60\,mm total length (\cref{design}). An advantage of using nitinol is that it resists plastic deformation if the whisker is accidentally smashed against a surface.

The base of the wire is inserted and glued into a 3D printed structure (Stratasys Objet24 with VeroWhitePlus, accuracy
 $\approx 0.1$\,mm.). 
 The structure consists of an undulating track with a groove, into which an optical fiber is placed and glued. The trajectory of the groove is dictated by the 8\,mm minimum bend radius of the fiber. 
 The fiber is Corning Ultra glass fiber with an acrylate coating. Bragg Gratings are located along the fiber adjacent to the site where each wire is anchored, so that bending moments at the base of the wire produce bending strains in the corresponding plastic structure. These, in turn, result in tensile or compressive strains in the FBGs. Each whisker anchor has two FBGs to measure moments about local $x$ and $y$ axes at the whisker base (\cref{design}).  An additional strain-isolated FBG serves for temperature compensation. The FBGs all have different nominal wavelengths, between 1525 and 1565 nanometers and are sampled simultaneously at 2\,kHz by a Micron Optics sm130 optical sensing interrogator. The general optical sensing approach follows that in \cite{hill1997fiber}. Finite Element Analysis was used to determine the FBG placements for maximum strain and corresponding sensitivity to light forces on the whisker (\cref{fea}). The FEA model includes the plastic structure and the thin, but much stiffer, glass fiber. 
 For protection, the base plastic structure is encapsulated in a thin layer of Shore 20A silicone rubber; the silicone is soft enough that it has a negligible effect on the strains detected by the FBGs. 

\section{Whisker Sim-to-Real Learning}
\label{sec:whisker_sim_to_real}
Collecting large-scale data data on contact positions from the real world is a challenging and laborious task. We therefore created a digital whisker sensor in the MuJoCo simulator, which 
approximates the real whisker dynamics.
The real-world data includes FBG wavelength shifts and camera images of the deflected whiskers, including contact locations---but does not provide whisker base moments directly. Accordingly, as shown in \cref{overview}, 
we train a transformer decoder using pair data of contact positions and base moments. In this section, we first introduce the data collection process and the simulation model. Finally, we present a calibration system designed to convert bending strains and associated wavelength shifts to orthogonal bending moments at the base of the needle, for reconciliation with the results of simulation.

\subsection{Contact Localization Assumptions}
We begin by summarizing the assumptions made to limit the scope of the problem.
\begin{enumerate}
\item Objects that come into contact are immobile in the world reference frame, meaning that if the robot does not move, the contact position remains unchanged.
\item There is at most one contact point on a whisker with the environment at any time. For surfaces that are locally concave in the direction of whisker travel, this means that we only capture the proximal whisker/object contact.
\item Frictional forces along the whisker are negligible in terms of their effect on the sensor, which is generally valid due to the properties of the nitinol whisker material, especially in water.
\end{enumerate}

\begin{figure}[t]
      \centering
      \includegraphics[width=0.95\linewidth]{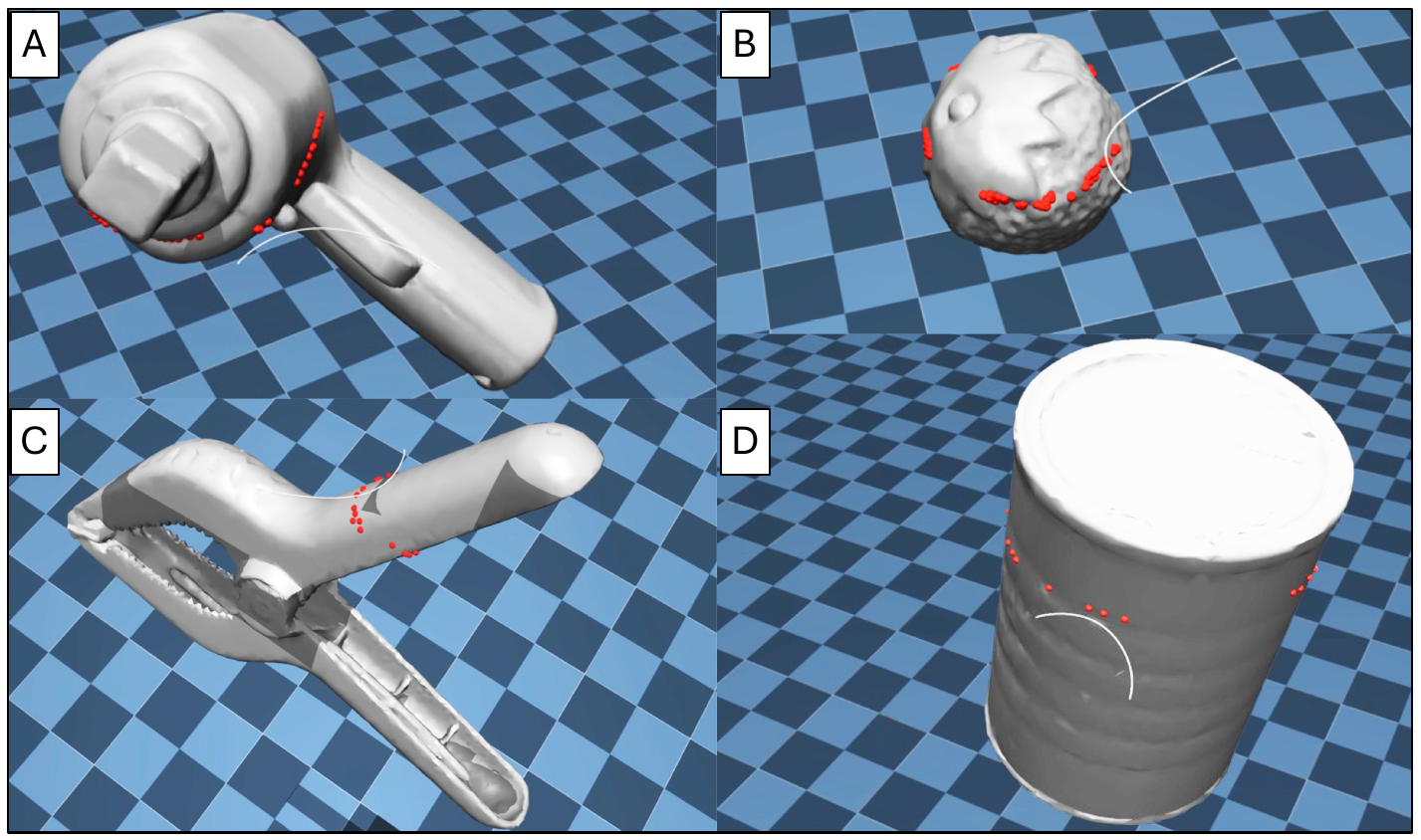}
      \caption{Data collection in MuJoCo of whiskers sweeping past (A) a power drill, (B) a strawberry, (C) a clamp, and (D) a coffee can in the YCB dataset. Red points indicate potential contact positions on the surface at the whisker sweeping plane.}
      \label{figurelabel}
\end{figure}

\subsection{Data Collection in Simulation}
\textbf{Whisker Sensor in MuJoCo} Our thin whiskers can be approximated as inextensible, elastic, one-dimensional objects in space. In MuJoCo, we simulate the whisker sensor using the Cable object \cite{mujoco2024}, which we represent as a sequence of capsules. Our aim is to capture the overall deflection of the whisker and the mapping from contact locations to base moments, without concern for higher order dynamics asasociated with whisker tip bouncing and vibrations. We therefore slightly increased the damping parameter for the capsule connections \cite{mujocoXMLReference} and reduced the surface impedance at contacts to reduce vibrations and base moment spikes arising from contact with rough surfaces and from sudden changes between free motion and contact \cite{mujocoDocumentation}.

In particular, the dimensionless damping parameter was incrementally adjusted from $\approx 0$ (consistent with a metal wire) to 0.03 in steps of 0.005. Additionally, to ensure simulation stability, we introduced a small rotational joint inertia of $3 \times 10^{-5} kg\cdot m^2$ as recommended in \cite{DeepMindMuJoCoArmature2023}. 

In addition, we reduced the contact impedance value $d$ and employed a higher-order impedance curve. This allowed us to balance the contact duration between minimal and maximal impedance values, $d_0$ and $d_{\text{max}}$, ensuring a gradual response without excessive penetration into objects. We conducted a grid search, varying the width from 0.001 to 0.010 with steps of 0.001, and exploring $d_0$ and $d_{\text{max}}$ from 0 to 1 with steps of 0.05. Details on how to modify these parameters can be found in the MuJoCo documentation \cite{mujocoTipsAndTricks}. As a result of these adjustments, the whisker bounces back at roughly half the speed of a bare nitinol wire when contacting a rigid object, and the collected data becomes more stable.

\begin{figure}[b!]
      \centering
      \includegraphics[width=0.9\linewidth]{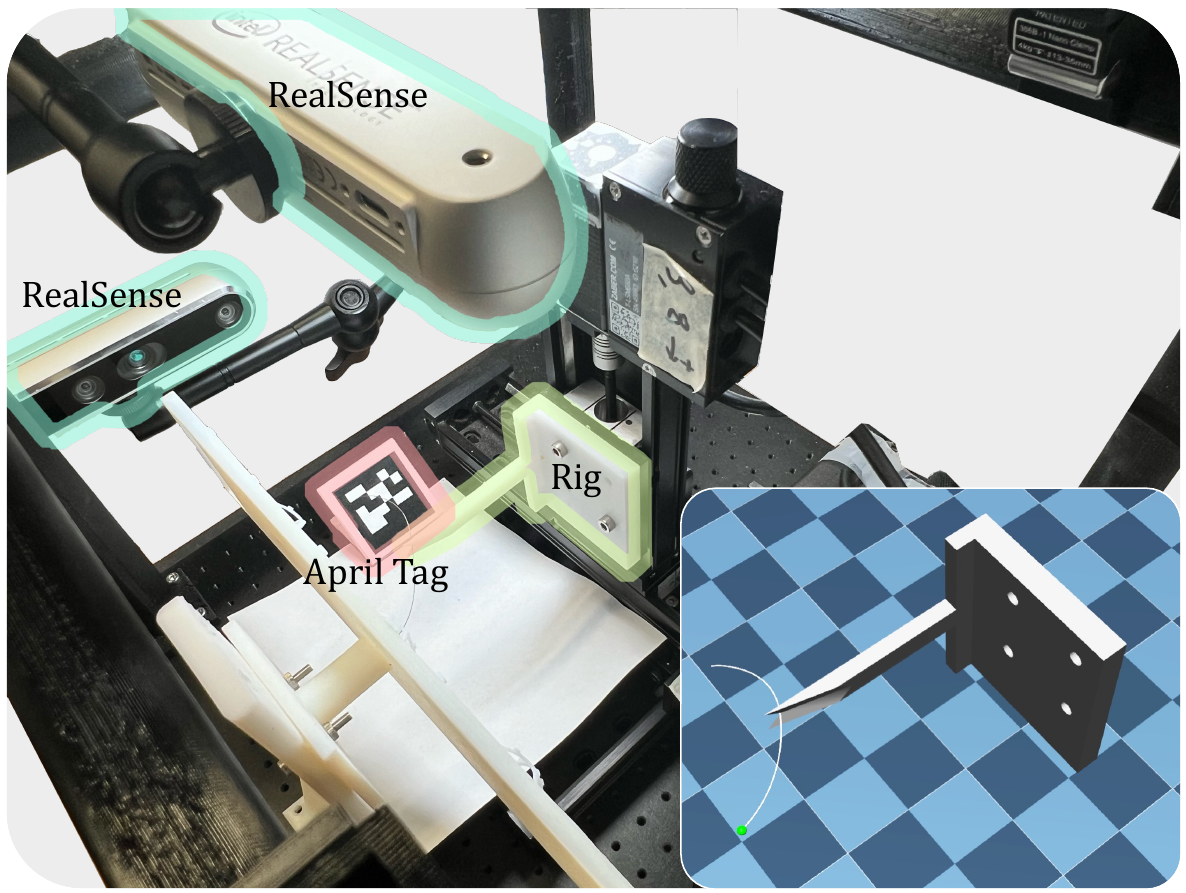}
      \caption{Real2Sim calibration setup in real world and simulation. Real world setup consists of two cameras, an April tag, and a calibration rig.}
      \label{cali}
\end{figure}

\textbf{Object Initialization} 
Since MuJoCo does not support the direct import of concave objects, it is necessary to decompose objects into convex components. We use CoACD \cite{wei2022approximate} for decomposition because it efficiently preserves the collision properties of the original shape without greatly increasing the number of components, enabling more precise and efficient object interactions in downstream applications. However, one limitation of this method is that when applied to meshes from scanned objects such as those in the YCB data set \cite{calli2015ycb},
it can remove surface details such as small bumps, grooves, and sharp corners. As will be discussed in the experiment section, this leads to the assumption that our experiments only support continuous contact. We will further discuss extensions to overcome this limitations in the future work section.

Building a whisker-sweeping dataset requires both quantity and quality, particularly in the diversity of objects swept by the whisker. To achieve this, we apply random rotations and transformations to the YCB objects, allowing the whisker sensor to interact with different faces and orientations. Specifically, after rotating the object, a random plane is selected, and the distance from the whisker sensor to the nearest point on the object is randomly set within a specified range. Given the curved shape of the whisker and the direction of travel, we assume that a single contact occurs with the object. If multiple contacts are detected, we select the one closest to the base of the whisker. Additionally, random acceleration is applied to the whisker sensor during sweeping to better simulate real-world conditions.

\subsection{WhiskerNet: Contact Prediction from the Torque}
Contact prediction from the sensor signal can be viewed as a sequence prediction problem. We designed a transformer decoder to learn the relationship between the contact and all the signals prior to that time step. We only select the torque signal in the y and z directions, as there is negligible torque about the whisker axis. Given the dataset \(\{\mathcal{S}, \mathcal{C}\}\), where \(\mathcal{S}\) represents the signals of the whisker base torque and \(\mathcal{C}\) represents the contact position in the whisker base coordinate, WhiskerNet infers the contact position from sequences of signals. Intuitively, it learns from the changes in the signals; for example, a change near the base can cause a large torque change. For each signal sequence \(\textbf{s} = [s_1, s_2, \ldots, s_n]\) and the contact position \(\textbf{c} = [c_1, c_2, \ldots, c_n]\), we learn the following distribution:

\begin{equation}
    p(\textbf{c} | \textbf{s}) = p(c_i | s_{<i}).
\end{equation}

Specifically, we encode the signal sequence \(\textbf{s}\) into the feature \(\textbf{f}^{\,0}\) via multiple linear layers and add the position embeddings. Then, several layers of transformer decoder layers attend to each other to learn the relationship features
\(\textbf{f}^{\,i}\) among them. A linear head maps the contact position \(\textbf{c}\) from the final features \(\textbf{f}^{\,n}\). The Mean Squared Error Loss for the predicted $\mathbf{c'}$ compared to the contact ground truth $\mathbf{c}$ is given by:

\begin{equation}
\mathcal{L}(\mathbf{c}, \mathbf{c'}) = \frac{1}{n} \sum_{i=1}^{n} (c_i - c_i')^2.
\end{equation}

\subsection{Real2Sim Calibration}
Due to structural issues, the FBG signals of the two axes in the real world setup are somewhat coupled. Therefore, a calibration mapping is required to map from the base torques in the simulation to the wavelength shifts. We gather the same calibration data from the real world and simulation. To obtain data we use a calibration rig attached to a multi-axis optical calibration stage. The rig includes a V-shaped groove to enforce a known contact location on the whisker. We sweep several linear trajectories to collect different contact positions and their corresponding sensor signals. The same calibration rig and calibration trajectories are used in the simulation in MuJoCo.

From the sim-to-real calibration, we obtain a dataset \(\mathcal{D} = \{ (s^{real}_i, s^{sim}_i) | i = 1, 2, \ldots, m \}\), where \(s_i^{real}\) is the wavelength of an FBG sensor, and \(s_i^{sim}\) is a torque on the whisker base in the simulation. We use Gaussian Process Regression (GPR) to map the real-world wavelength \(s^{real}\) to simulation torque \(s^{sim}\). In GPR, the kernel function defines the covariance between pairs of random variables. We choose a Thin-plate (TP) kernel for its robust performance in the mapping, as noted in previous work with a different whisker sensor \cite{lin2024navigation}. The kernel function is defined as:
\begin{equation}
    k_{TP} = \frac{(2|r|^2 - 2Rr_2 + R^3)}{12}
\end{equation}
where \(R\) is the radius of the region within which the TP model will minimize the second-order gradient. Before mapping the real-world data using GPR, a series of heuristic methods first filter the signal noise and filter out the signals below a certain threshold.

\begin{figure}[b]
      \centering
      \includegraphics[width=0.9\linewidth]{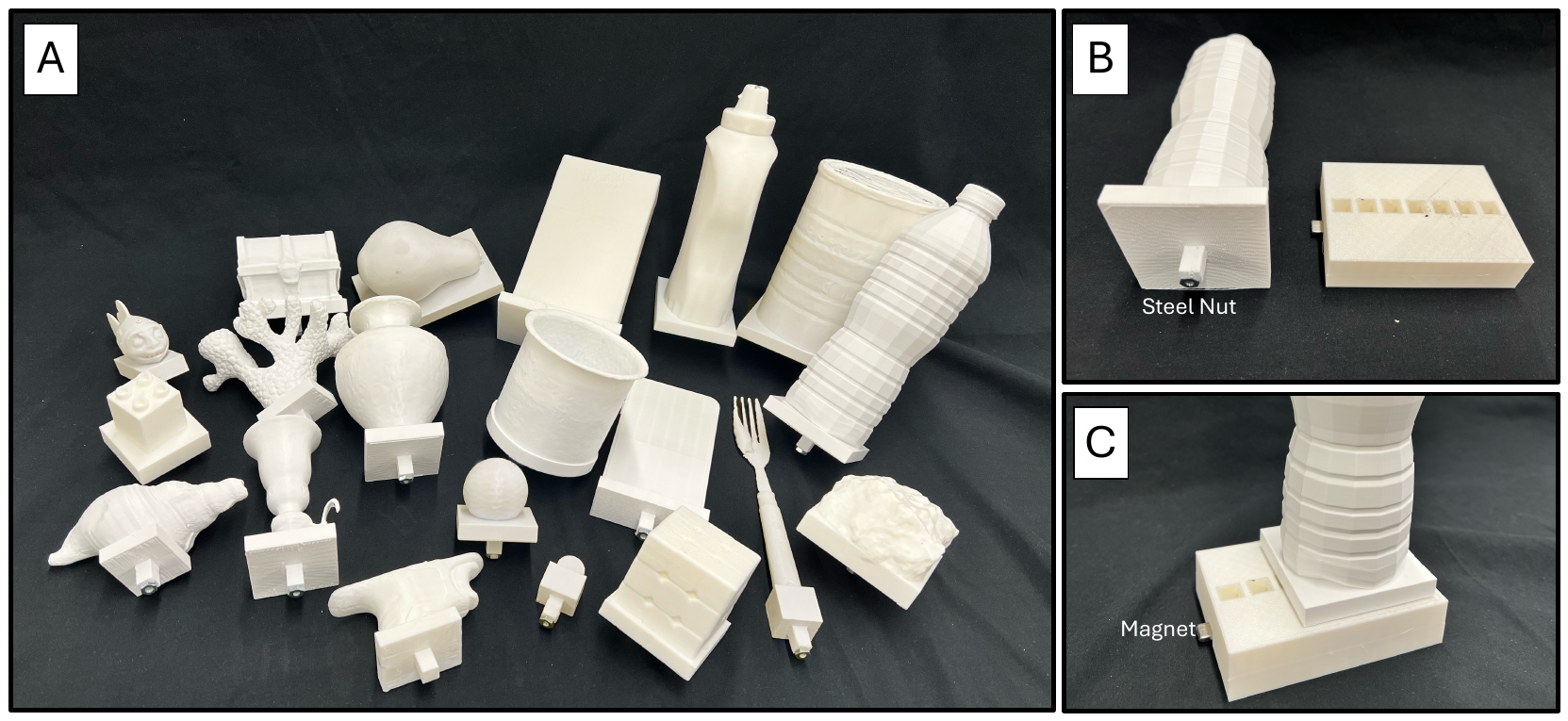}
      \caption{(A) Twenty 3D-printed objects were used in experiments, each with (B) a steel nut at the bottom for attachment (C) to a magnet.}
      \label{holder}
\end{figure}

\section{Experiment Setup}
We performed experiments both in simulation and in the real world, including an underwater experiment, a speed ablation study, and a sim2real ablation study. This section provides details about both setups, the sim2real calibration setup, and the WhiskerNet implementation details.

\begin{table*}[t]
\centering
\caption{Simulation and Real World Experiment Results }
\label{tab:compare-table} 
\resizebox{\textwidth}{!}{%
\begin{tabular}{>{\centering\arraybackslash}m{2cm}cccccccccc}
\toprule
RMSE errors (unit: mm) 
& \includegraphics[width=0.8cm,height=1.2cm]{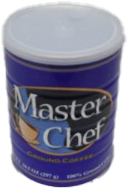} 
& \includegraphics[width=0.95cm,height=0.95cm]{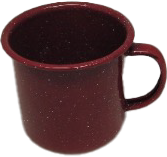} 
& \includegraphics[width=0.95cm,height=0.95cm]{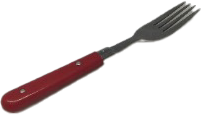} 
& \includegraphics[width=0.6cm,height=0.6cm]{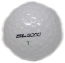} 
& \includegraphics[width=0.8cm,height=1.2cm]{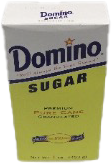} 
& \includegraphics[width=0.85cm,height=0.95cm]{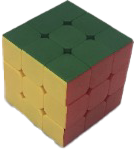} 
& \includegraphics[width=0.6cm,height=0.8cm]{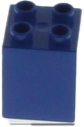} 
& \includegraphics[width=0.7cm,height=1.2cm]{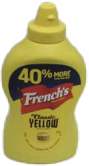} 
& \includegraphics[width=0.95cm,height=0.95cm]{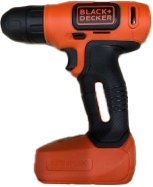} 
& \includegraphics[width=1cm,height=0.7cm]{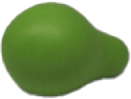} \\
\midrule
Simulation & 0.32 & 0.39 & 0.59 & 0.22 & 0.2 & 0.27 & 0.47 & 0.38 & 0.29 & 0.36 \\
Real World & 1.10 & 0.85 & 1.37 & 0.94 & 1.34 & 0.83 & 1.43 & 1.51 & 1.06 & 1.91 \\
\midrule
& \includegraphics[width=0.65cm,height=0.95cm]{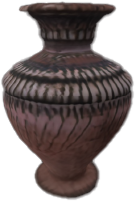} 
& \includegraphics[width=0.95cm,height=0.8cm]{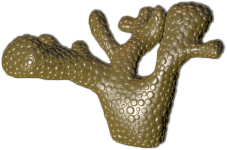} 
& \includegraphics[width=0.95cm,height=0.7cm]{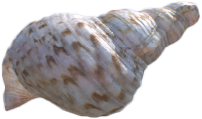} 
& \includegraphics[width=0.5cm,height=1.2cm]{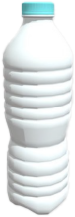} 
& \includegraphics[width=0.55cm,height=1.2cm]{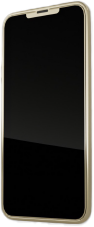} 
& \includegraphics[width=0.95cm,height=0.95cm]{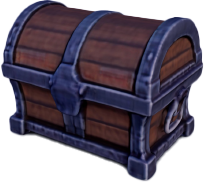} 
& \includegraphics[width=0.95cm,height=0.95cm]{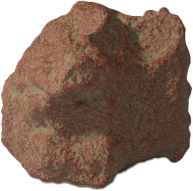} 
& \includegraphics[width=0.5cm,height=0.6cm]{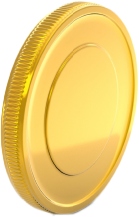} 
& \includegraphics[width=0.95cm,height=0.65cm]{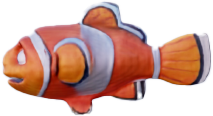} 
& \includegraphics[width=0.6cm,height=0.95cm]{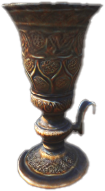} \\
\midrule
Simulation & 0.15 & 0.22 & 0.51 & 0.56 & 0.56 & 0.56 & 0.22 & 0.91 & 1.38 & 0.35 \\
Real World & 0.88 & 1.18 & 1.38 & 1.64 & 1.09 & 1.24 & 1.89 & 2.71 & 3.09 & 2.01 \\
\bottomrule
\end{tabular}%
}
\end{table*}

\subsection{Simulation Setup}
To train and evaluate WhiskerNet in the simulation environment, we built our digital twin in MuJoCo using a dual camera setup to obtain the whisker shape in the real world and collect a large-scale dataset.
 We used 78 objects from YCB dataset of which mesh files are available \cite{calli2017yale}. For each object, we collected 200 samples of sweeping with random orientations. The initial moving speed of whiskers was randomly set between 3 and 7\,mm/s and we applied a changing acceleration of $\pm \ 0.4 \ mm^2/s$.

During data collection, We record the torque at the whisker base and the corresponding contact positions. We filter out signals with torque values exceeding a certain threshold and those with contact jumping problems, defined by the difference between the current and previous contact positions. After compiling the dataset, we downsample the data to 5\,Hz. To augment the data, we introduce random no-contact signals at the beginning of contact and truncate the signals to a predetermined maximum length. The dataset is then randomly split into 
80\% for the training set and 20\% for the validation set. For testing, we select the same objects used in the real-world experiments. The performance is evaluated using the Root Mean Square Error (RMSE) between the predicted contact positions and the actual object surface.

\subsection{Immersed Experiment Setup}
\begin{figure}[pb!]
      \centering
      \includegraphics[width=0.9\linewidth]{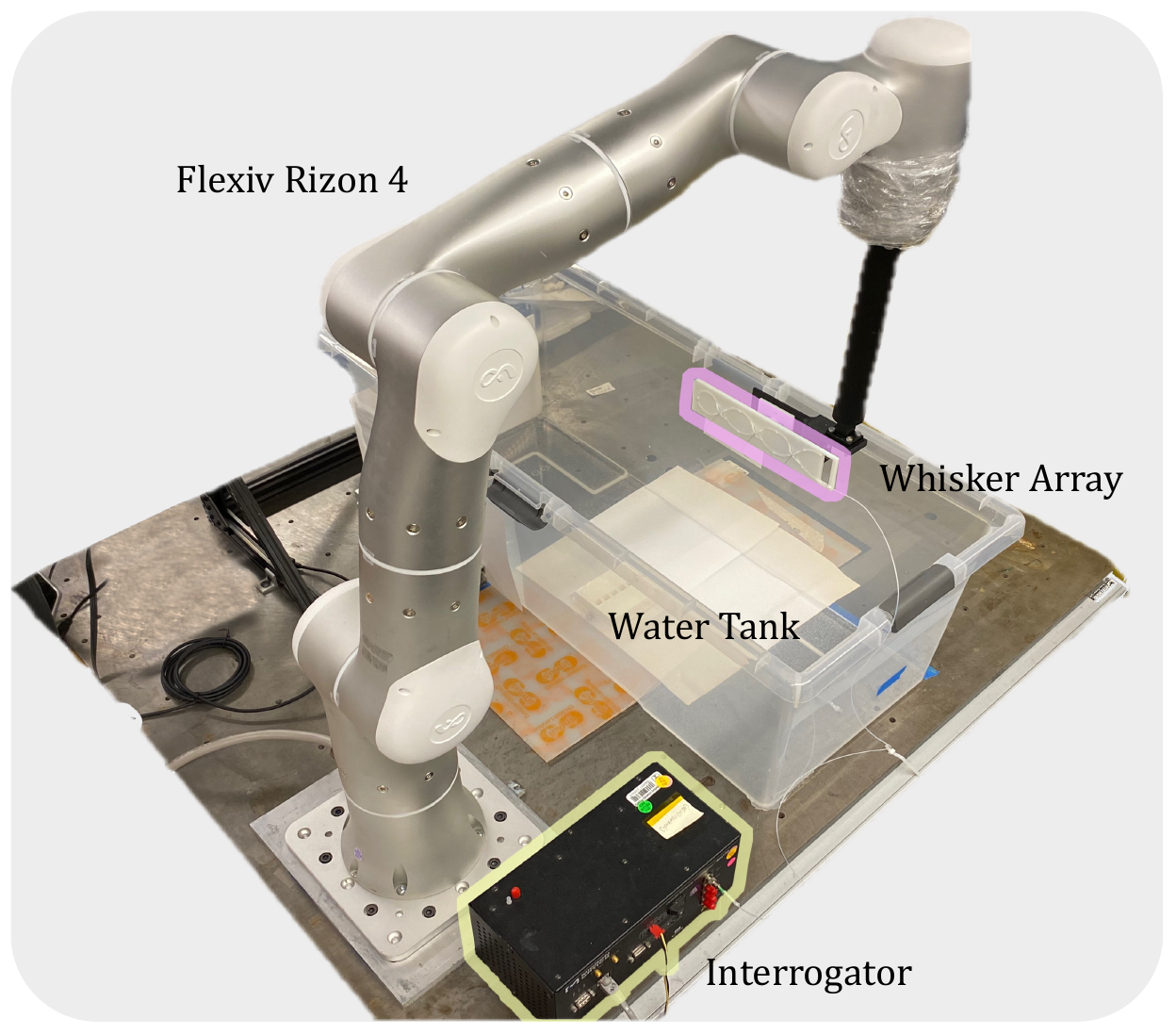}
      \caption{Setup of underwater experiments: whisker array is mounted on a robot arm with a 3D printed extension to keep the arm dry as whiskers pass over objects. Optical fiber is connected to an interrogator outside the tank.}
      \label{setup}
\end{figure}

We conducted underwater experiments to validate the accuracy of the whisker sensor, as illustrated in \cref{setup}. The whisker sensor is mounted on a Flexiv Rizon 4 robot arm using a 3D-printed connector. Sensor signals are transmitted through a Micron Optics sm130 optical sensing interrogator. For testing, we selected 10 seen objects from the YCB dataset and 10 unseen objects commonly found in underwater environments. The seen objects are chosen to represent a variety of geometries and categories within the YCB dataset, while the unseen objects are selected to include more challenging cases. As shown in \cref{holder}, to mount all the test objects, we add a pillar on the bottom of the objects and design a holder with seven cube-shaped slots to secure them. A steel nut attached to the end of each pillar is magnetically fixed to the holder, ensuring the objects remain stable against any bounce forces and allowing for easy attachment and removal.

Prior to testing, the whisker array is fully immersed in water for 30 minutes to stabilize the signal in response to temperature changes. The robot arm then moves from right to left, sweeping whiskers across the objects. The sweeping height is individually selected for each object to ensure continuous and convex contact in a single trial. Examples include the handle of a power drill, the bottom portion of a piece of artificial coral, and the convex part of a shell (\cref{tab:compare-table}).  After each trial, the object is rotated 90 degrees, with a total of four trials conducted per object. 

The underwater experiment and sim2real ablation study use the same setup, with the exception that the in-air test in the sim2real ablation study is conducted without water in the tank. In the speed ablation study, we select the same coffee can as the testing object and sweep on its first side. For the underwater experiment and sim-to-real ablation study, a moving speed of 4\,mm/s is used, with random variations between \(80\%\) and \(120\%\) at each time step. For speed ablation study, we select the coffee can as the test object. A video demonstration with three views is included in the supplementary materials. The performance is evaluated using the RMSE between the predicted contact positions and the object surfaces.

\subsection{Sim-to-real Calibration Setup}
To map the wavelength signal to the simulation torque, we built matching calibration setups in hardware and in simulation \cref{cali}. The real world calibration is performed with a Zaber model LSM100A linear optical stage and a 3D-printed calibration rig. We ran the calibration in both simulation and real-world environments using the 9 same linear trajectories. For each trajectory, we evenly collected values at three points. Therefore, we collect 27 pairs of wavelength and torque values in total, which cover the sensing range of whiskers. 

\subsection{WhiskerNet Implementation Details} 
For the WhiskerNet, we use a Transformer decoder to map the signal to the contact position. The hidden size of the signal encoder is 64. The Transformer decoder consists of 6 layers with causal attention and 8 attention heads. The hidden size is 128 and the FFN hidden size is 512. 
During training, the batch size is 64 and the learning rate is \(3 \times 10^{-4}\). The dropout rate is set to 0.1. We use the Adam optimizer \cite{kingma2014adam} with \(\beta_1=0.9\), \(\beta_2=0.999\), and \(\epsilon=1 \times 10^{-8}\).

\begin{table}[t!]
\centering
\caption{Speed Ablation Study Results}
\label{tab:speed}
\begin{tabular}{>{\centering\arraybackslash}m{2cm}cc}
\toprule
(unit: mm)
& w/o random & w/ random \\
\midrule
4 mm/s & 0.42 & 0.88\\
6 mm/s & 0.88 & 1.35\\
8 mm/s & 1.36 & 2.13\\
10 mm/s & 1.74 & 2.32\\
12 mm/s & 2.25 & 2.99\\
\bottomrule
\end{tabular}%
\end{table}

\section{Results}
We present the results of our underwater experiment and two ablation studies to evaluate our sim2real contact prediction performance. 

\subsection{Main Results}
Table \ref{tab:compare-table} presents the accuracy of our whisker sensor in both simulated and real-world underwater environments within a water tank. In the simulation, the accuracy for all objects is high, with errors within 1.5\,mm. For real world experiments on the in-domain YCB objects,the model effectively captures the geometry of cylindrical objects with minimal curvature changes (row 1, columns 1-4). It also performs well on objects with flat surfaces (row 1, columns 5-7), except in cases where the object is positioned close to the whisker, causing sharp corners, like those of the LEGO block and the sugar box, to drag the whisker and hinder its movement. These types of objects perform well overall, as the contact point remains stable without lateral slipping on the contact surface. Beyond standard shapes like cylinders or cubes, the model also shows good performance on more complex shapes (row 1, columns 8-9). However, the accuracy is slightly reduced when slipping happens in the direction orthogonal to whisker travel (row 1, column 10).

In real-world experiments with unseen objects, the model continues to demonstrate robust performance on items with standard or simple geometries, such as a vase, bottle, and chest. Even with highly complex surfaces, like a rock, the accuracy remains within 2 mm. This can be attributed to the fact that, although the object itself may be out of distribution, its geometric characteristics are not. However, there are some scenarios where our model encounters difficulties. These include very flat and thin objects (row 2, column 8), situations where jumping occurs, and concave objects with discontinuous contact. Detailed analysis of these cases is provided in the following case studies.

\subsection{Speed Ablation Study}
Table \ref{tab:speed} presents the results of our speed ablation study, demonstrating a clear correlation between whisker movement speed and model performance. As the whisker’s speed increases, the model’s performance gradually declines. Notably, when the speed exceeds 8\,mm/s, which lies beyond the range of our training data, there is a marked deterioration in performance. Additionally, introducing variability in the whisker speed exacerbates prediction errors compared to a consistent speed scenario.

\begin{table}[H]
\centering
\caption{Sim2Real Study Results}
\label{tab:sim2real-table}
\begin{tabular}{>{\centering\arraybackslash}m{2cm}ccc}
\toprule
(unit: mm) & Simulation & In air & In water \\
\midrule
Vase & 0.15 & 0.85 & 0.88 \\
Power Drill & 0.29 & 0.91 & 1.06 \\
Rock & 0.22 & 1.90 & 1.89 \\
\bottomrule
\end{tabular}%
\end{table}

\subsection{Performance in Simulation and Real-World Environments}
We evaluated the performance of our whisker sensor in both simulation and real-world environments, specifically in air and underwater conditions. The sensor was tested on three diverse and representative objects selected from 20 testing objects. In simulation, all objects demonstrate high accuracy, with errors of less than 1\,mm. The performance in air is comparable to that in water, which can be attributed to our design choice of using a 0.3\,mm thick nitinol wire and our low speeds so that drag forces from the water are negligible.

\subsection{Case Study of Successes and Failures}
\begin{figure}[t]
      \centering
      \includegraphics[width=0.85\linewidth]{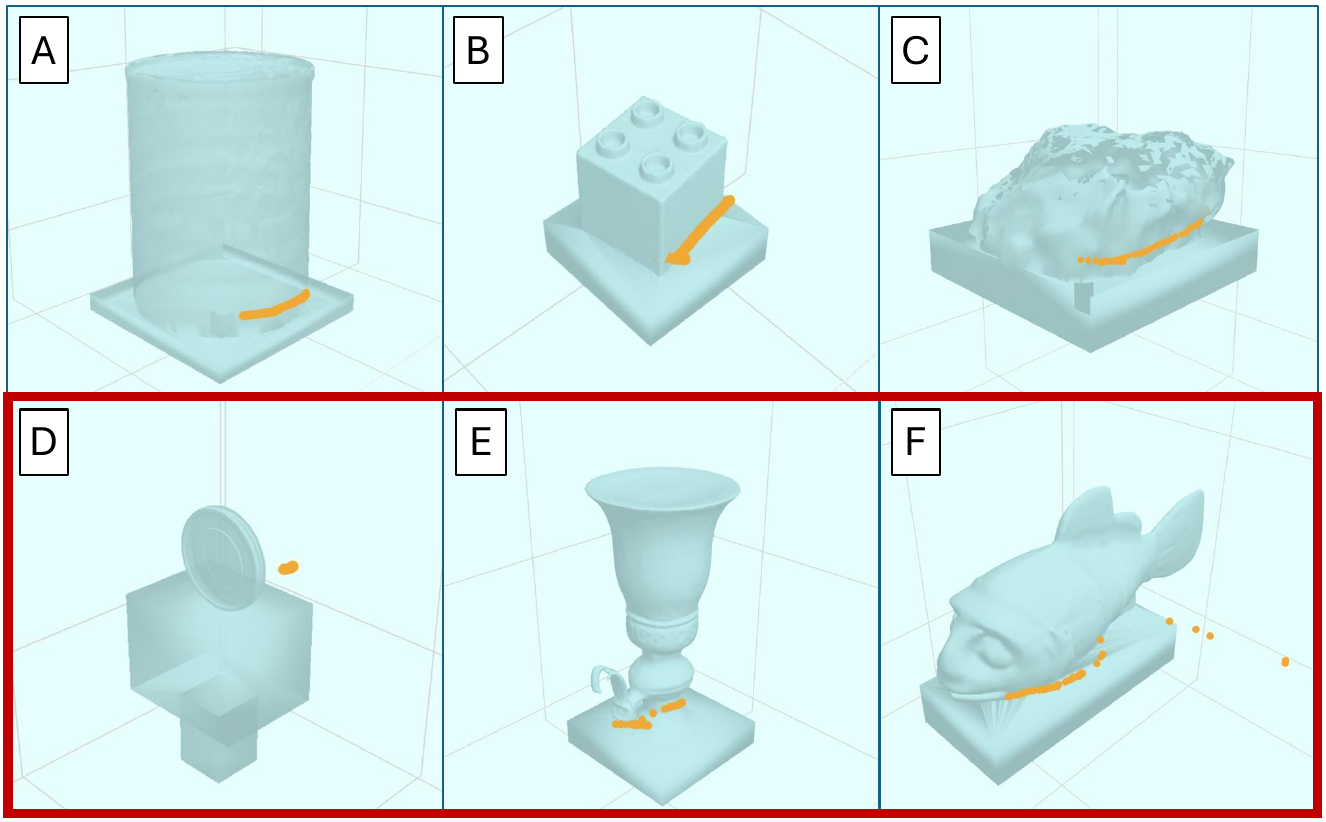}
      \caption{Visualization of successful contact position predictions in underwater environment on (A) a coffee can, (B) a LEGO block, and (C) a rock. The figure also shows examples of prediction failures, such as on (D) a narrow coin, (E) a concave lamp, and (F) a complex fish.}
      \label{casestudy}
\end{figure}

As shown in Fig. \ref{casestudy}, the three objects at the top represent successful cases, while the three objects at the bottom illustrate failure cases. For the coffee can (A), the model performs best on smooth surfaces, with the whisker signal spiking when sweeping across the object’s corner. For the Lego block (B), the model performs well on the two faces but produces an outlier at the corner. The model also works well on uneven objects like the rock (C). Although it cannot capture the fine-grained details of the surface, it successfully predicts its shape.

One of the failure patterns observed occurs with thin and flat objects, such as the coin (D). In our training dataset, the contact prediction trajectories occur on continuous surfaces rather than being stuck at corners or edges. As a result, the model incorrectly predicts a slight movement in the contact position from its previous point, leading to significant errors. Another failure case occurs when the whisker interacts with concave objects, such as the lamp (E). When the whisker brushes against the gap between the handle and the body of the lamp, it inevitably makes multiple contacts, leading to inaccurate predictions in the gap. The third failure case occurs when the whisker jumps or slips on the surface, as seen with the fish (F). The contact position is lost when there is insufficient friction to keep the whisker in continuous contact with the surface. As discussed below, these all represent cases associated with discontinuous whisker contact tracing.

\section{CONCLUSIONS AND FUTURE WORK}

Underwater manipulation in cluttered and confined environments remains a challenge due to the limited sensing capabilities of current technologies. To enhance robotic perception underwater, our design features a low-force, non-intrusive whisker system based on FBG technology, capable of accurately predicting contact positions using a single optical fiber. This design is robust, waterproof, and well-suited for underwater environments. Additionally, we propose a sim-to-real learning framework that involves collecting large-scale data in a digital twin environment and training our deep learning WhiskerNet model to predict contact locations using signal history. The framework also includes a calibration method to map the signals and transfer the trained model to real-world scenarios. This approach allows us to release a constraint from our previous work \cite{lin2022whisker}, where accurate robot proprioception is required. Our results demonstrate significant success in transferring simulations to real-world applications, both in air and in water.

For future work, we plan to address the limitations of our current method. Specifically, we aim to improve our perception algorithms by accounting for multiple contacts and discontinuous contacts. Our current approach assumes that only one contact occurs at a time and contacts are continuous along the trajectory. Thus, we filter out samples with multiple or discontinuous contacts in both our training and testing datasets. However, these types of interactions are common, particularly when dealing with complex object geometries. In response, we plan to develop methods for identifying and handling such cases, including segmenting the data to determine when a new object interaction begins. We also intend to expand our method to accommodate a broader range of material properties and surface geometries. Currently, we assume friction is negligible and use only 3D-printed objects for testing. We aim to deepen our understanding of whisker interactions with different objects. Additionally, we plan to increase the whisker’s movement speed to support more realistic manipulation scenarios and refine the data collection process by integrating simulation with real-world fine-tuning. Moreover, we recognize the need to develop a more precise metric for evaluating the accuracy of our contact predictions.

Furthermore, we are interested in exploring the sensor’s potential for water flow detection. The high sampling rate of FBG signals suggests that whiskers have significant potential in sensing water flow dynamics, which could open new avenues for underwater sensing and manipulation tasks. To achieve this, we will explore new methods to distinguish between water flow information and contact signals.

\section*{ACKNOWLEDGMENT}
We are grateful for support from the Zhulong Innovation Fellowship and the Stanford ME SURI program, with additional support from TRI Global.

\bibliographystyle{ieeetran}
\bibliography{ref}
\end{document}